\documentclass[runningheads]{llncs}
\usepackage[T1]{fontenc}
\usepackage{graphicx}
\usepackage{caption}
\usepackage{subcaption}
\usepackage{booktabs}
\usepackage{amssymb}

\newcommand{\ProjName}{SAIBench}

\usepackage{url}

\usepackage{breakurl}

\begin{document}
\title{Does AI for science need another ImageNet Or totally different benchmarks? A case study of machine learning force fields}
 \author{Yatao Li\inst{1,2,3} \and
 Wanling Gao\inst{1} \and
 Lei Wang\inst{1} \and
 Lixin Sun\inst{3} \and
 Zun Wang\inst{3} \and
 Jianfeng Zhan\inst{1,2}}
 
 \institute{Institute of Computing Technology Chinese Academy of Science, No.6 Kexueyuan South Road, Haidian Beijing 100190, China
 \email{\{gaowanling,wanglei\_2011,zhanjianfeng\}@ict.ac.cn} \and 
 University of Chinese Academy of Sciences, No.19(A) Yuquan Road, Shijingshan Beijing 100049, China \and 
 Microsoft Research, No. 5 Dan Ling Street, Haidian Beijing 100080, China
 \email{\{yatli,lixinsun,zunwang\}@microsoft.com}
 }

\maketitle

\begin{abstract}

AI for science (AI4S) is an emerging research field that aims to enhance the accuracy and speed of scientific computing tasks using machine learning methods. 
Traditional AI benchmarking methods struggle to adapt to the unique challenges posed by AI4S because they assume data in training, testing, and future real-world queries are independent and identically distributed, while AI4S workloads anticipate out-of-distribution problem instances.
This paper investigates the need for a novel approach to effectively benchmark AI for science, using the machine learning force field (MLFF) as a case study.
MLFF is a method to accelerate molecular dynamics (MD) simulation with low computational cost and high accuracy.
We identify various missed opportunities in scientifically meaningful benchmarking and propose solutions to evaluate MLFF models, specifically in the aspects of sample efficiency, time domain sensitivity, and cross-dataset generalization capabilities.
By setting up the problem instantiation similar to the actual scientific applications, more meaningful performance metrics from the benchmark can be achieved.
This suite of metrics has demonstrated a better ability to assess a model's performance in real-world scientific applications, in contrast to traditional AI benchmarking methodologies.
This work is a component of the \ProjName ~project, an AI4S benchmarking suite. The project homepage is~\url{https://www.computercouncil.org/SAIBench}.

\end{abstract}

\section{Introduction}

Benchmarks are extensively utilized in computer science research and the IT industry to assess and compare the performance patterns of various types of entities, from abstract and mathematically specified problem definitions and algorithms to fully materialized software + hardware systems \cite{bischlASlibBenchmarkLibrary2016,wangBigDataBenchBigData2014}. 
The term ``benchmark'' originated from land measurement practices, where marks were carved onto a stone, creating a ``fixture'' for mounting measurement equipment.
Modern computer science benchmarks follow a similar concept. 
To measure performance metrics, a ``fixture'' is created by instantiating the target problem into a standardized set of computing resources and tasks. 
For instance, in machine learning benchmarks, problem instantiation is achieved by mapping high-level goals (e.g., image classification) to a concrete dataset, such as ImageNet \cite{dengImageNetLargescaleHierarchical2009}.
A critical factor in benchmarking is to ensure that the problem instantiation aligns with the stakeholders' interests.
In the context of machine learning benchmarks, this means that the chosen dataset should cover all typical scenarios implied by the high-level goals. 
Consequently, if a machine learning model performs well on the dataset, it is expected to perform well in real-world applications.

AI for science (AI4S) is an emerging research field that focuses on leveraging machine learning methods to improve accuracy and speed in scientific computing tasks \cite{argonnenationallaboratoryAIScienceReport,zhangArtificialIntelligenceScience2023}.
Benchmarking AI4S is crucial, as it allows scientific researchers to evaluate the quality of an AI4S machine learning model and ensure its successful integration into the scientific computing pipeline.

The ultimate goal of AI4S is to assist scientific researchers in exploring the unknown, which often challenges fundamental assumptions in machine learning. 
For instance, traditional machine learning assumes that the training and testing instances share the same distribution, and so do instances from any future queries. 
While this assumption works well with traditional workloads such as object recognition in ImageNet, where the dataset is indeed randomly sampled from all possible objects ``in the wild'', a scientific computing pipeline is well expected to encounter entirely new instances.
In other words, the ``in-distribution'' assumption fails in AI4S scenarios, where encountering ``out-of-distribution'' data is anticipated. 

Consequently, traditional AI benchmarking methods struggle to adapt to the AI4S context due to this ``out-of-distribution'' challenge.
Good performance in training and simple testing no longer guarantees that the model will perform well when integrated into a real-world scientific computing pipeline.
Adhering to conventional AI benchmarking practices will invariably result in a biased problem instantiation that is misaligned with the objectives of AI4S. 

This ponders the question: do we need to carry over the practices from conventional AI benchmarking by collecting a comprehensive dataset like ImageNet, or do we need a completely different approach to benchmark AI for science effectively?
In this paper, we study a particular example in AI for science, molecular dynamics (MD) simulations.
MD simulation serves as a crucial tool that is extensively employed in chemical physics, materials science, biophysics, and related fields.
MD simulation models the motion of atoms and molecules within a chemical system and how it evolves over time.
Machine learning was proposed to accelerate MD simulation.
At the core of the machine learning acceleration lies the machine learning force field (MLFF), which computes the forces applied to each atom in the chemical system.

We identify numerous missed opportunities in scientifically meaningful benchmarking MLFF models, and we propose solutions that allow us to investigate the behavior of MLFF models in greater detail. Our contribution is as follows. 
1. We propose an evaluation of the sample efficiency of MLFF models, specifically focusing on their performance in scenarios with sparse data. This is in contrast to conventional AI workloads, such as large-scale language models and image recognition tasks, which often have access to vast amounts of data.
2. While conventional AI benchmarks assume that the samples in the dataset are independent and identically distributed, we propose to take advantage of the fact that the MD simulation produces time-series data,  and we furthermore evaluate the time domain sensitivity of the models.
3. Contrasting to conventional AI benchmarks that typically treat different datasets as separate entities, we propose the development of cross-dataset generalization tests for MLFF models. 
4. While our primary objective is to evaluate the performance of MLFF models, we also uncover an intriguing correlation between the test results and a similarity metric known as Smooth Overlap of Atomic Positions (SOAP). This discovery can, in turn, help us to improve the simulation pipeline.

\section{Preliminaries}

MD simulation is an essential tool that simulates atomic motions within chemical systems, providing key insights for computational chemistry, biology, and physics to unravel thermodynamic and kinetic phenomena.
The key of an MD simulation is to integrate atomic motion by applying computed forces to each atom and subsequently displacing atoms following Newton's Second law $\vec{f}_i = m_i\vec{a}_i$, where $\vec{f}_i$, $m_i$, $a_i$ are the force, mass, and acceleration of atom $i$.
Traditionally, the atomic forces are computed with empirical inter-atomic potentials or with \emph{ab initio} methods as the negative gradient of potential energy (Eq. \ref{eq.forces}).
\begin{equation}
    \vec{f}_i = - \frac{\partial}{\partial \vec{x}_i} E(\vec{x}_1, \vec{x}_2, \dots, \vec{x}_n)
    \label{eq.forces}
\end{equation}
In empirical potentials, the potential energy functionals are relatively simple analytical equations, such as in Lennard-Jones potential \cite{jonesDeterminationMolecularFields1997}.
They often assume that each atom is only affected by its neighboring atoms,
\begin{equation}
    \vec{f}_i = - \frac{\partial}{\partial \vec{x}_i} E_i(\vec{x}_{i}, \vec{x}_{j_1}, \vec{x}_{j_2}, \dots, \vec{x}_{j_{n_i}})
    \label{eq.forces-emperical}
\end{equation}
where $j_k$ are atoms that are close to the atom $i$ with $|\vec{x}_{j_k} - \vec{x}_i| < r_\mathrm{cut}$.
This locality assumption leads to an $O(N_\mathrm{atom})$ complexity for classical force fields.
On the other hand, the \emph{ab initio} methods solve Schr\"odinger's equation to obtain potential energy, with a complexity of $O(N_\mathrm{electron}^K), ~K = 3-5$.
Even though the empirical potentials are easy to compute and often scale linearly with the number of atoms $N_\mathrm{atom}$, they are limited to systems without the formation and breaking of chemical bonds. 
In contrast, \emph{ab initio} MD (AIMD) methods incorporate quantum mechanic effects, providing a more precise representation of the potential in chemical reactions, albeit at the expense of prohibitive computation time.
This accuracy-cost trade-off makes it prohibitive to use MD simulation to model systems with chemical reactions.

Machine learning enables a new solution, MLFF, to solve this accuracy-cost trade-off problem.
It uses neural networks \cite{behlerGeneralizedNeuralNetworkRepresentation2007} or Gaussian Process \cite{bartokGaussianApproximationPotentials2010} for the potential energy functional.
The functional is more complicated than empirical force fields but keeps the $O(N_\mathrm{atom})$ scaling.
Hence MLFF can scale up to large molecular systems that were previously impossible to simulate with empirical force fields while maintaining an acceptable level of accuracy \cite{jiaPushingLimitMolecular2020}.

As mentioned before, MD simulation is an iterative process where each step computes the forces applied to each atom in the system. 
Based on the force field, the algorithm iteratively updates the velocities and positions of the atoms over time steps, generating a trajectory of the molecular system that illustrates the evolution of the system over time.
Given a chemical system of $N_\mathrm{atom}$ atoms, the force field takes the cartesian coordinates and atomic numbers as input and outputs the potential energy and forces as written in Eq. \ref{eq.forces} 

The construction and training of MLFF constitute the following steps.
\begin{enumerate}
    \item Obtain training data from \emph{ab initio} methods, which contains cartesian coordinate $\vec{x}_1, \vec{x}_2, \dots, \vec{x}_N$, atomic numbers $z_1, z_2, \dots, z_N$, total potential $E$, and forces $\vec{f}_1, \vec{f}_2, \dots, \vec{f}_N$.
    For simplicity, all the variables can be written as matrices: $\mathbf{x}, \mathbf{F} \in \mathbb{R}^{N \times 3}$, $\mathbf{Z} \in \mathbb{R}^{N \times 3}$, $E \in \mathbb{R}$, where $\mathbf{F} = -\frac{\partial}{\partial \mathbf{X}} E$.
    \item construct a machine learning model 
\begin{equation}
    \widetilde{E}, \widetilde{\mathbf{F}} = \mathcal{F}(\mathbf{X}, \mathbf{z})
\end{equation}
where $\mathcal{F}$ can be a neural network or a Gaussian Process model.
    \item Design a loss function $\mathcal{L}$ over both the forces and the energy. An example is to use mean square error.
    \begin{equation}
    \mathcal{L} = \sum_{\mathrm{batch}} [ \alpha (E - \widetilde{E})^2 + \beta \sum_i \sum_{u=x, y, z}(f_{i, u}-\widetilde{f}_{i, u})^2 ]
    \end{equation}
    \item Optimize the functional $\mathcal{F}$ over the training data, to minimize $\mathcal{L}$.
    \item Evaluate the performance of the trained model.
\end{enumerate}

\section{Related Works}

The rapid development of MLFF models \cite{batatiaMACEHigherOrder2022,musaelianLearningLocalEquivariant2023,wangViSNetScalableAccurate2022} has garnered significant interest in various applications, such as drug discovery and material design. These applications require high numerical accuracy from MD simulations, making the quality of a trained MLFF model crucial for successful deployment. As highlighted by \cite{fuForcesAreNot2022,wangImprovingMachineLearning2023}, a suboptimal MLFF model can introduce errors in the predicted potential energy and forces, which accumulate throughout the iterative simulation steps. These accumulated errors can result in simulation failures, manifesting as non-physical conformations of the chemical system or numerical runaway conditions.

Numerous benchmarking suites have been developed to evaluate machine learning workloads \cite{gaoAIBenchScalableComprehensive2019,gaoAIBenchScenarioScenarioDistilling2021,mattsonMLPerfTrainingBenchmark}. 
The evaluation metrics chosen by each benchmark correspond to the interests of the stakeholders. For instance, machine learning model benchmarks emphasize convergence rate and accuracy and focus on metrics such as the validation and test loss function performance, while industrial AI benchmarks prioritize cost-efficient model deployment \cite{gaoAIBenchIndustryStandard2019} and concentrate on performance metrics such as model throughput and hardware resource utilization. 
Recently, benchmarks specifically targeting AI for science are also proposed \cite{thiyagalingamScientificMachineLearning2022}, which carries over the methodologies of conventional AI benchmarking.

In the context of MLFF, the primary interest of stakeholders lies in the successful integration of the MLFF model into MD simulations. However, conventional benchmarking methods encounter difficulties in accurately reflecting this objective. 
Traditional AI evaluation metrics primarily concentrate on statistical performance across the entire dataset; however, Tong et al. \cite{wangImprovingMachineLearning2023} observed that failures in MD simulations with MLFFcan likely be traced back to a few poor force predictions, resulting in irrecoverable error accumulation.
Moreover, conventional AI evaluation metrics are derived directly from the difference between model output and ground truth data, while Tian et al. \cite{fuForcesAreNot2022} pointed out that stability in simulation does not align with training and testing performance.
Therefore, we argue that conventional benchmarking methods are not well suited for evaluating MLFF.

Despite the efforts in existing works, it remains desirable to characterize MLFF better. This motivates us to develop a novel benchmark specifically for MLFF, aiming to ensure MLFF quality. 

\section{Benchmarking MLFF}\label{sec:methodology}

The benchmark uses focuses on evaluating NequIP \cite{batznerEquivariantGraphNeural2022}, an equivariant graph neural network architecture specifically designed for learning MLFF from ab-initio calculations. NequIP emerges as a prominent representative of the state-of-the-art in this field, distinguished by its exceptional data efficiency and superior performance when compared to previous HDNN-style neural networks \cite{behlerGeneralizedNeuralNetworkRepresentation2007} and kernel-based methods \cite{bartokGaussianApproximationPotentials2010}. NequIP's remarkable data efficiency enables accurate modeling of MLFF with minimal training data, making it an appealing choice for scenarios where data availability is limited or costly to obtain.

To assess the performance of the machine learning model, we utilize a benchmarking fixture created from the revised MD17 (rMD17) dataset \cite{christensenRoleGradientsMachine2020}. The original MD17 dataset \cite{chmielaMachineLearningAccurate2017} comprises data points obtained from MD simulation trajectories based on density functional theory (DFT), encompassing a predefined set of molecules. The rMD17 dataset further enhances the MD17 dataset by employing a more accurate level of theory, thereby mitigating numerical noise and improving data quality.
It is important to call out that we adopt a different fixture setup compared to conventional machine-learning-style practices. In a conventional machine-learning-style setup, it would assume that the data points (from both the MD17 and rMD17 datasets) are randomly sampled from the ground truth problem space. Training and testing are subsequently conducted by randomly partitioning the data into subsets. 
However, MD17 data points are drawn from simulated trajectories, resulting in inherent correlations in the time domain. Consequently, randomly sampling training and test subsets can lead to the interleaving of data points from different time steps. While this scenario aligns with the ideal situation in MLFF-powered MD, where simulated data covers a wide range of molecule conformation space, we argue that it is not the case for the MD17/rMD17 dataset, which will be demonstrated in the forthcoming experimental results.
The rMD17 dataset is often consumed in a random train/test split manner, as mentioned before because the data points are not ordered as a trajectory time series. This can be mitigated by sorting the data with the ``old\_index'' field, which maps the data points back to the original MD17 and restores temporal order in the data. Our benchmarking fixture is established on this calibrated dataset by splitting out the last 10\% data in the time series as the test subset.

\subsection{Sample Efficiency}

\begin{figure}[t!]\centering
   \begin{subfigure}{0.48\textwidth}
     \centering
     \includegraphics[width=\linewidth]{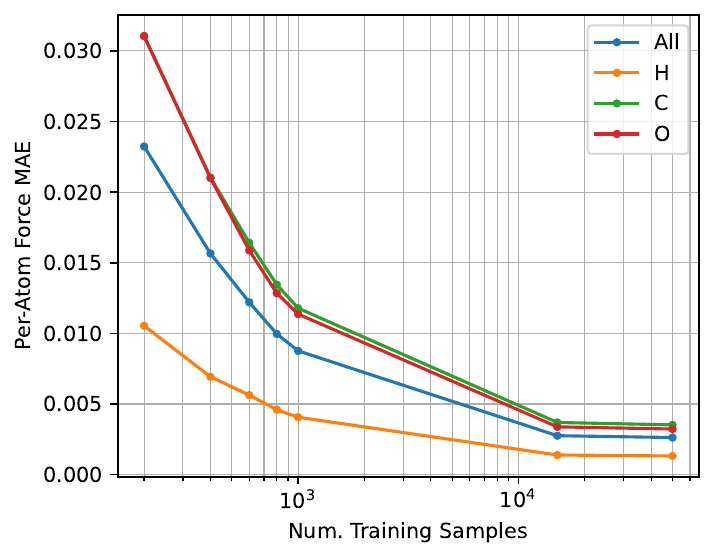}
     \caption{Training data size vs. Forces MAE}\label{fig:1a}
   \end{subfigure}
   \hfill
   \begin{subfigure}{0.48\textwidth}
     \centering
     \includegraphics[width=\linewidth]{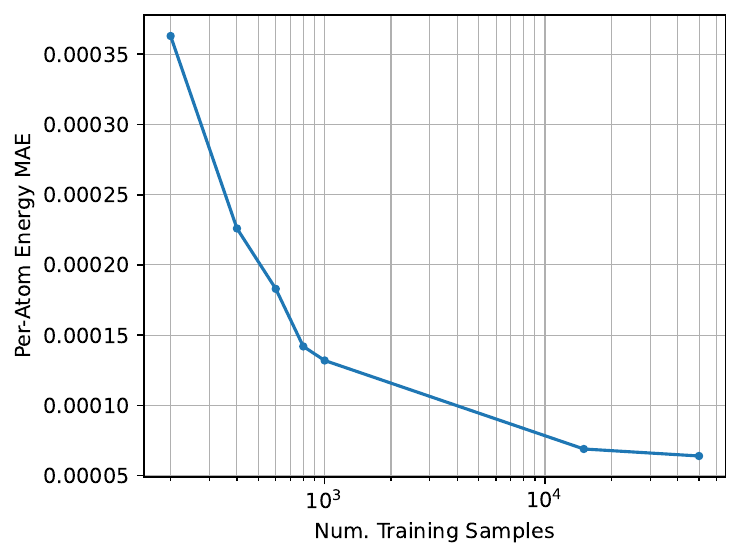}
     \caption{Training data size vs. Energy MAE}\label{fig:1b}
   \end{subfigure}
   \caption{Sample Efficiency Benchmarks}
   \label{fig:1}
\end{figure}

We evaluate the sample efficiency of the model by fixing the training data window to the first 90\% of the trajectory simulated on an aspirin molecule and progressively sample more data (200, 400, 600, 800, 1000, 15000, and 50000 samples, respectively) from the window into different training subsets, and compare the performance of trained models on the test subset. The training process for each subset is given a fixed wall time budget, allowing all to converge properly. We compare both per-atom force mean average error (MAE) and per-atom energy MAE for the trained models. 
The benchmarking results are illustrated in figure~\ref{fig:1}.

We can see that the model has good sample efficiency, achieving per-atom energy MAE of less than 4 meV over the test data given only 200 training samples. More specifically, given a fixed training data window, the performance of the trained models progressively improves with more training data points. Both the performance on energy and forces follow a similar trend, where increasing the number of samples results in proportional improvements up to 1000 samples, but the gain decreases exponentially afterward, where the benefit of increasing the size of the training set from 1K to 15K samples is not as good as increasing from 200 to 400, albeit at the cost of much longer computational cost spent in each training epoch. 

In conventional AI benchmarking, it is not practical to evaluate the performance on parts of each data point, such as the first ten tokens generated from a model or the accuracy of prediction about the top-left part of an image. However, the structural and composable nature of molecular data allows for a more versatile projection of performance results, providing a unique opportunity to evaluate AI performance in multiple dimensions and giving more insights into the model's behavior and capabilities.
For example, Figure~\ref{fig:1a} additionally presents per-atom forces MAE for each species of atoms (Hydrogen, Carbon, and Oxygen). The analysis reveals that the error on different species generally follows the same trend, and the error on Hydrogen is significantly lower than the others due to its low atomic charge. Interestingly, the errors are not strictly proportional to the atomic charge of each species, as one might expect the errors of Oxygen to be proportionally higher than that of Carbon, but the data shows otherwise.

This observation suggests that force prediction is sensitive to the structural configuration of the molecule in addition to the invariant features of each atom. Moreover, it indicates that the model captures more structural information to reflect the steepness in the potential energy surface than an empirical potential energy equation. 

\subsection{Time-series Extrapolation}
The previous benchmark evaluates the model performance when the entire range of trajectory up to the test window is available to the training process. That is, the model is trained on data sampled from 9 times more time steps (90\%) to predict the immediately upcoming steps (10\%). 
In real-world MLFF-powered MD simulations, it is expected that the MLFF model should be able to support longer runs with more time steps, where the training window might not cover a large number of time steps compared to the inference steps and may not be immediately adjacent to the inference window.
To evaluate the model's performance under such conditions, multiple variations of benchmarks are created using a grid-scan method to vary the size of the training window and its starting point. The training window sizes are set to 30\%, 45\%, 60\%, 75\%, and 90\% of the whole trajectory, while the starting points are set to 0\%, 15\%, 30\%, 45\%, and 60\% of the whole trajectory. For each of these training window variants, two models are trained with 1K and 15K data points sampled from the window, respectively, and their performance is tested on the final 10\% of the trajectory.

\begin{figure}[t!]\centering
  \begin{subfigure}{1.0\textwidth}
    \centering
    \includegraphics[width=1.0\linewidth]{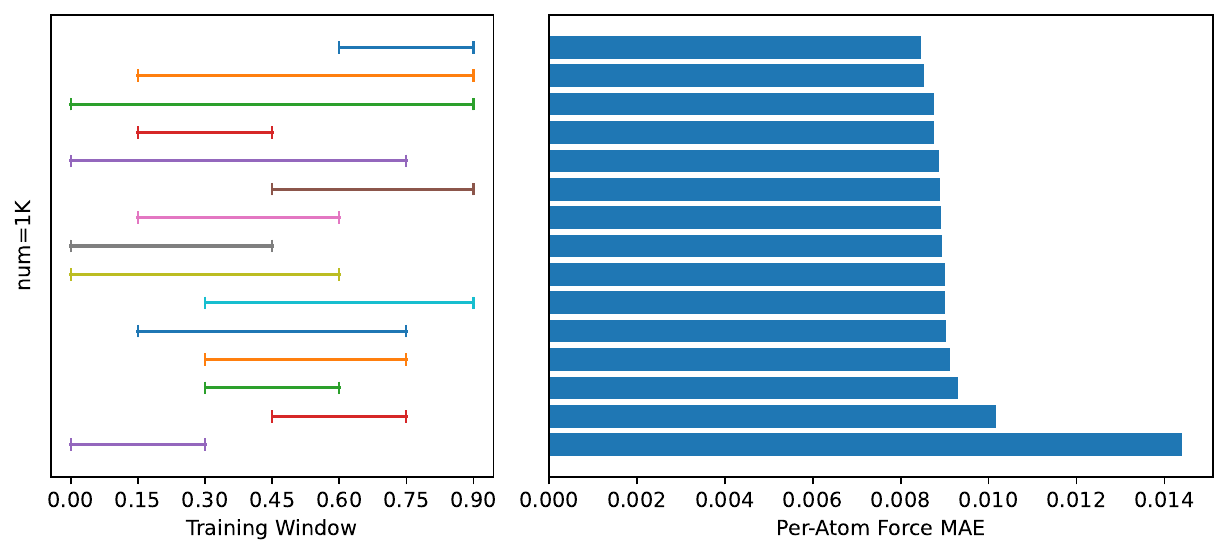}
    \caption{1K data points sampled from the window}
  \end{subfigure}
  \begin{subfigure}{1.0\textwidth}
    \centering
    \includegraphics[width=1.0\textwidth]{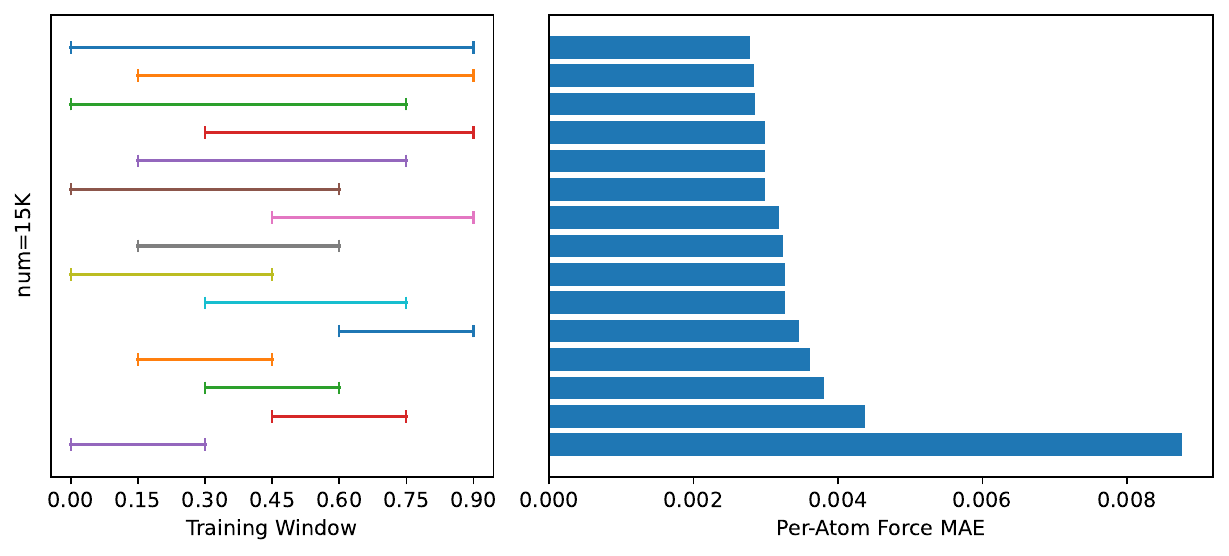}
    \caption{15K data points sampled from the window}
  \end{subfigure}
  \caption{Time-Series Extrapolation Benchmarks}
  \label{fig:2.1}
\end{figure}

Figure~\ref{fig:2.1} presents the time-series extrapolation benchmarking results. Each horizontal line segment in the left part of the chart represents a training window variant, with starting/ending points in the 0\%-90\% range. The bar on the right corresponds to the per-atom force MAE evaluated on the test window for the model trained on this specific training window. The data is sorted by test performance, with the training window on the first row having the best test performance.
The data shows that the test performance varies significantly with different training windows, and the patterns differ for 1K and 15K training samples. For the 1K samples, the best window is the one closest to the test window, with the narrowest range (60\%-90\%). In contrast, for the 15K samples, the best window is the widest (0\% to 90\%).
This observation suggests that, even though the model demonstrates excellent sample efficiency, 1K samples still lead to underfitting when a large window is used. The reason is that there is not enough data within each subsection of the window for the model to generalize to similar cases effectively.

Both the 1K and 15K charts show that a small window temporally distant from the test window (0\%-30\%) results in the worst performance. This observation suggests that maintaining the model's accuracy over a long trajectory is challenging, as it may not have enough information from distant data to generalize effectively to the test window.

\begin{figure}[t!]\centering
  \includegraphics[width=1.0\textwidth]{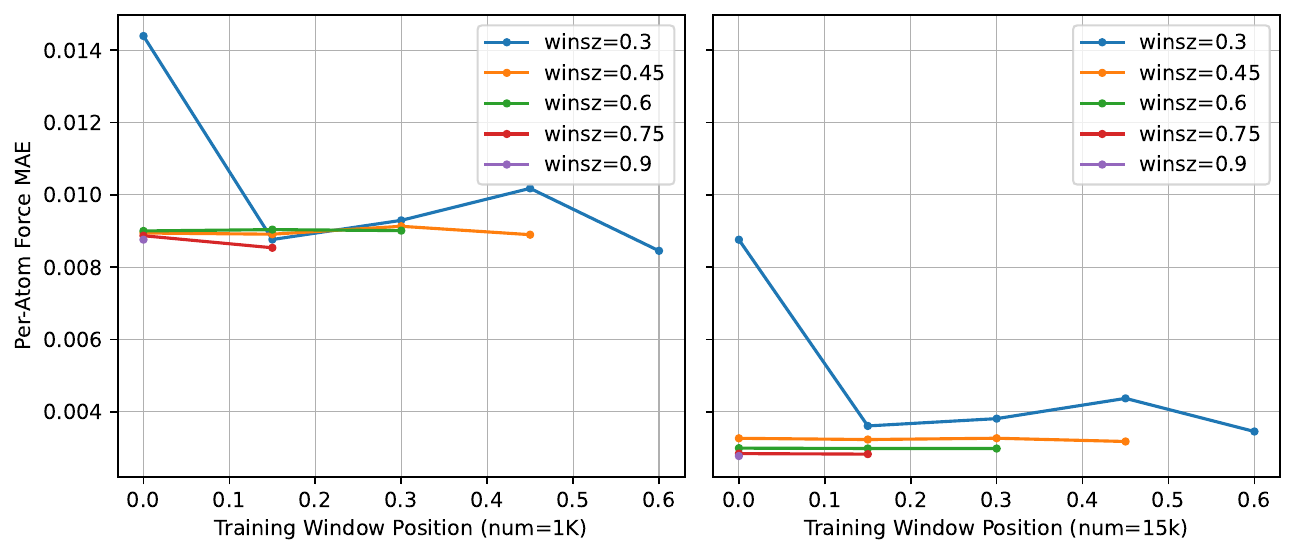}
  \includegraphics[width=1.0\textwidth]{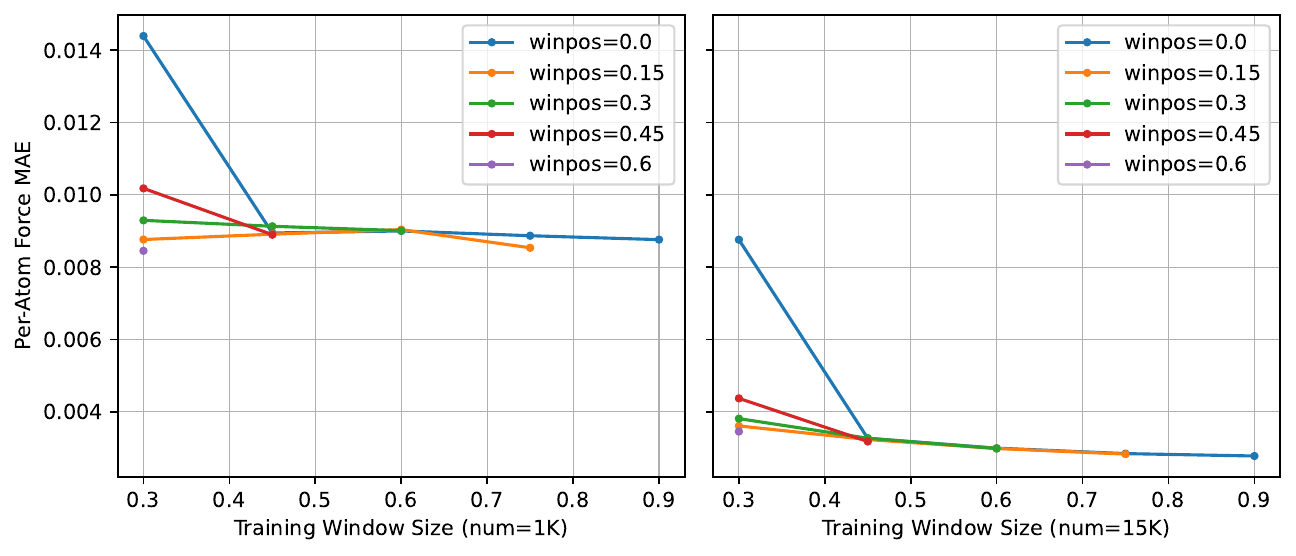}
  \caption{Time-Series Extrapolation Benchmarks (projected)}
  \vspace{-10px}
  \label{fig:2.2}
\end{figure}

\begin{figure}\centering
  \includegraphics[width=0.48\textwidth]{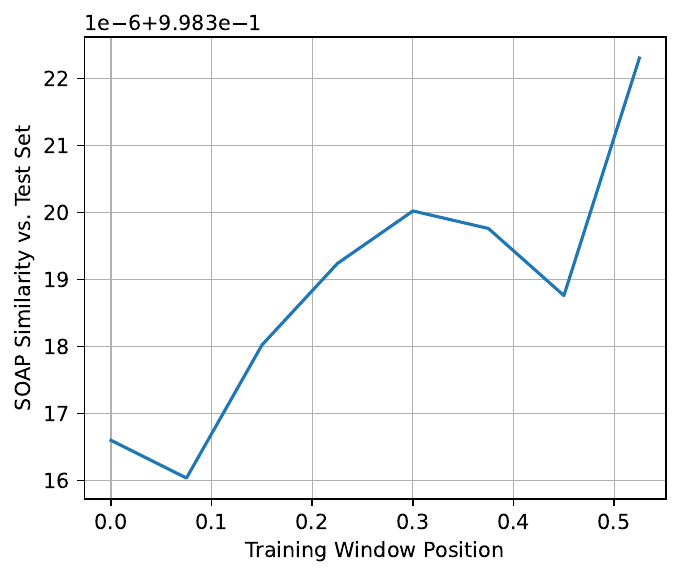}
  \caption{Training data window vs. Test window SOAP similarity}
  \label{fig:soap}
\end{figure}

In order to better evaluate the models trained on different windows, it is a good idea to project the results by grouping the data by window size and plotting each group to show the performance changes based on different window positions. Similarly, we can analyze how the performance changes with different window sizes for each starting position. This approach is visualized in figure~\ref{fig:2.2}.
From the data, it is observed that both 1K and 15K models exhibit a pattern where, given a fixed window starting position (in short, winpos), the performance increases monotonically with the window size, except for 1K samples with a winpos of $0.15$. However, given a fixed window size, the performance does not monotonically increase as the training window moves closer to the testing window.
To understand why this occurs, the SOAP (Smooth Overlap of Atomic Positions) descriptor \cite{bartokRepresentingChemicalEnvironments2013} is leveraged. The SOAP descriptor computes a high-dimensional feature vector for a given molecular system, allowing for the comparison of different molecular configurations. The correlation between two molecular configurations can be calculated by computing the cosine similarity of their corresponding SOAP descriptors.
By computing all pairwise correlations between the training windows with a 30\% range and the testing window, the mean average values are used to represent the similarity of the training windows to the test window. This information is visualized in figure~\ref{fig:soap}.

The similarity curve presented in the figure demonstrates that similarity does not monotonically increase as the training window moves closer to the testing window. This result has two significant implications.
First, the finding suggests that the trajectory does not constantly move away from the initial molecular configuration. Instead, it occasionally "bounces back" into the data distribution of earlier trajectories. This behavior indicates that the MD simulation may exhibit a certain degree of periodicity or recurring patterns in the molecular configurations, which can be essential in understanding the system's underlying behavior.
Second, the result shows a clear relationship between the window similarity metric and the test performance. This relationship suggests that a real-world MLFF-powered MD system could leverage this metric as an accuracy indicator. When the similarity drops below a certain threshold, it signals that the MLFF-powered MD loop is heading towards out-of-distribution space. In such cases, the model may require further fine-tuning to maintain accuracy and stability.

\subsection{Cross-Molecule Generalization Benchmarks}

The revised MD17 dataset consists of multiple MD simulation trajectories for a fixed set of molecules. Traditionally, separate benchmarking fixtures are created for different molecules because cross-molecule performance is poor and considered impractical for simulation purposes.

However, it is important to consider this as an opportunity to evaluate the out-of-distribution generalization capabilities of a target machine learning model. While the results may not be practical for direct simulation purposes, they can provide valuable insights into the relationship between the potential energy surfaces of different molecules and the fine local structures within these molecules.

\begin{table}
  \begin{center}
  
  \begin{subfigure}{0.24\textwidth}
    \centering
    \includegraphics[width=0.8\linewidth]{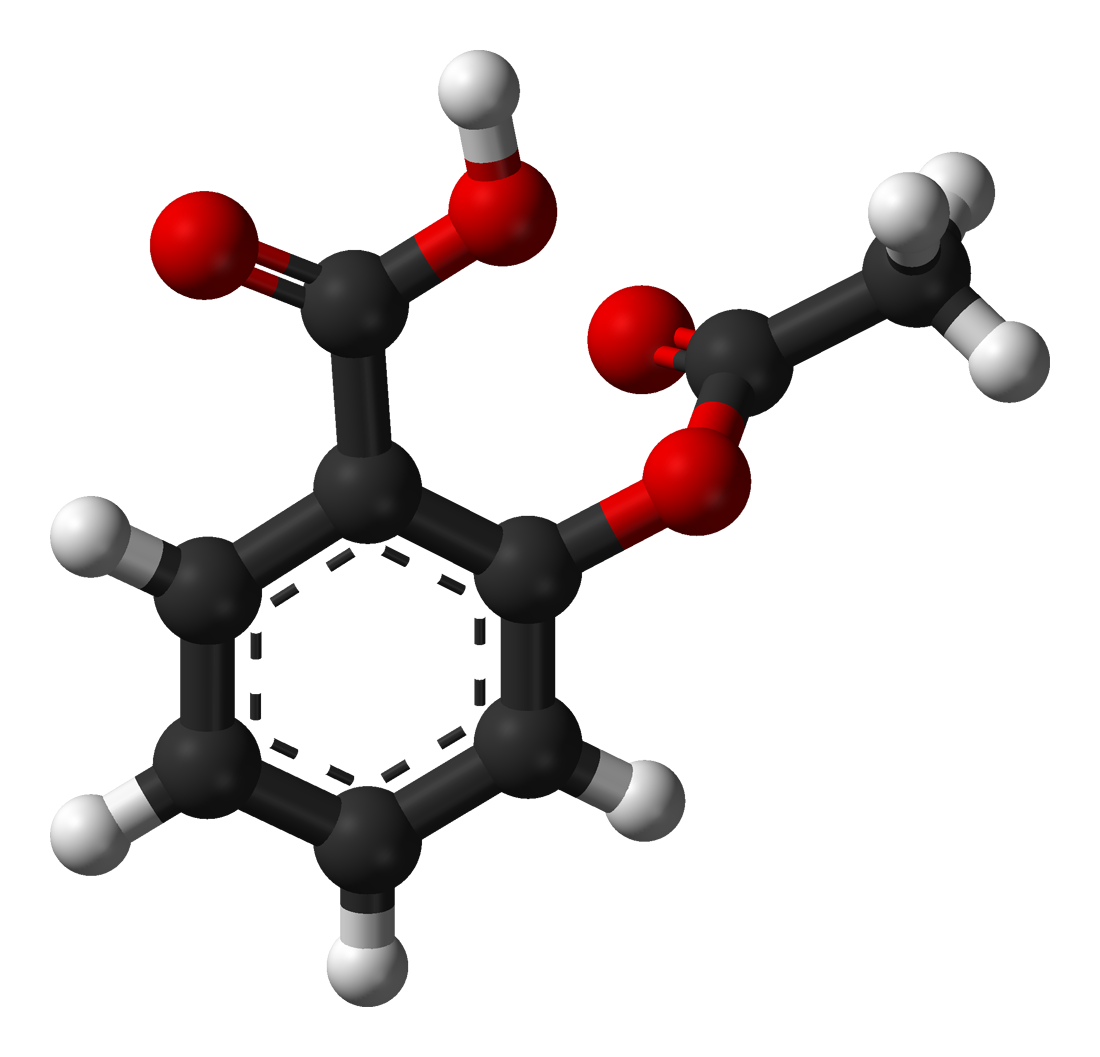}
    \caption{Aspirin}
  \end{subfigure}
  \begin{subfigure}{0.24\textwidth}
    \centering
    \includegraphics[width=0.8\linewidth]{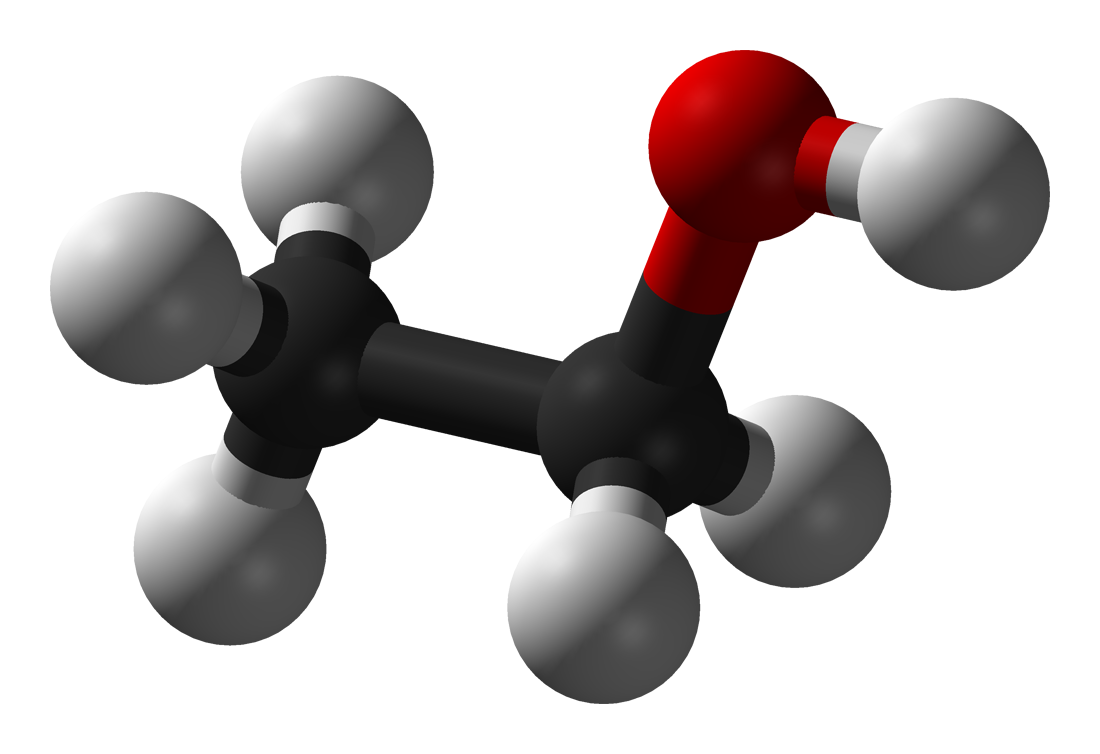}
    \caption{Ethanol}
  \end{subfigure}
  \begin{subfigure}{0.24\textwidth}
    \centering
    \includegraphics[width=0.8\linewidth]{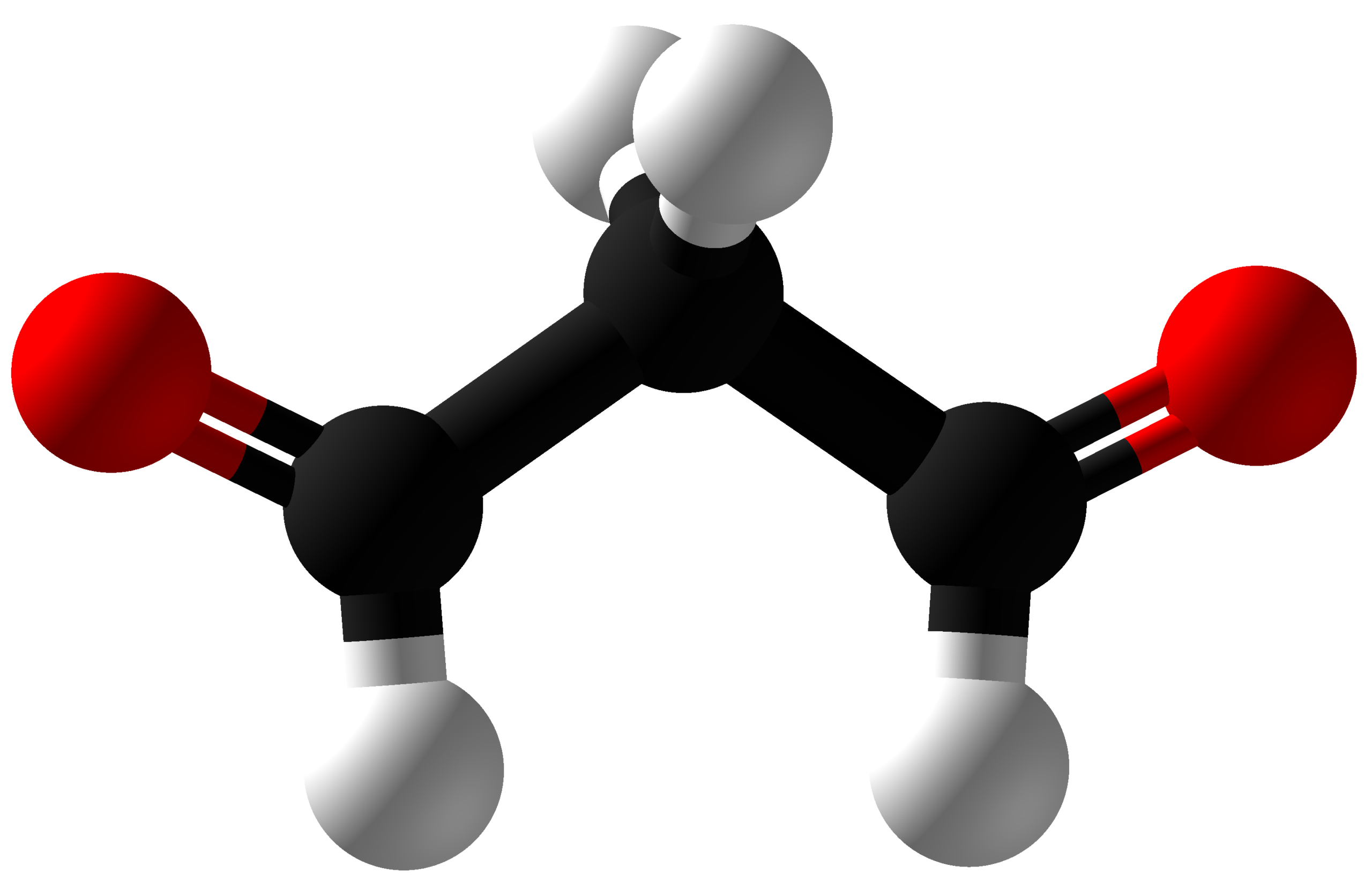}
    \caption{Malonaldehyde}
  \end{subfigure}
  \begin{subfigure}{0.24\textwidth}
    \centering
    \includegraphics[width=0.8\linewidth]{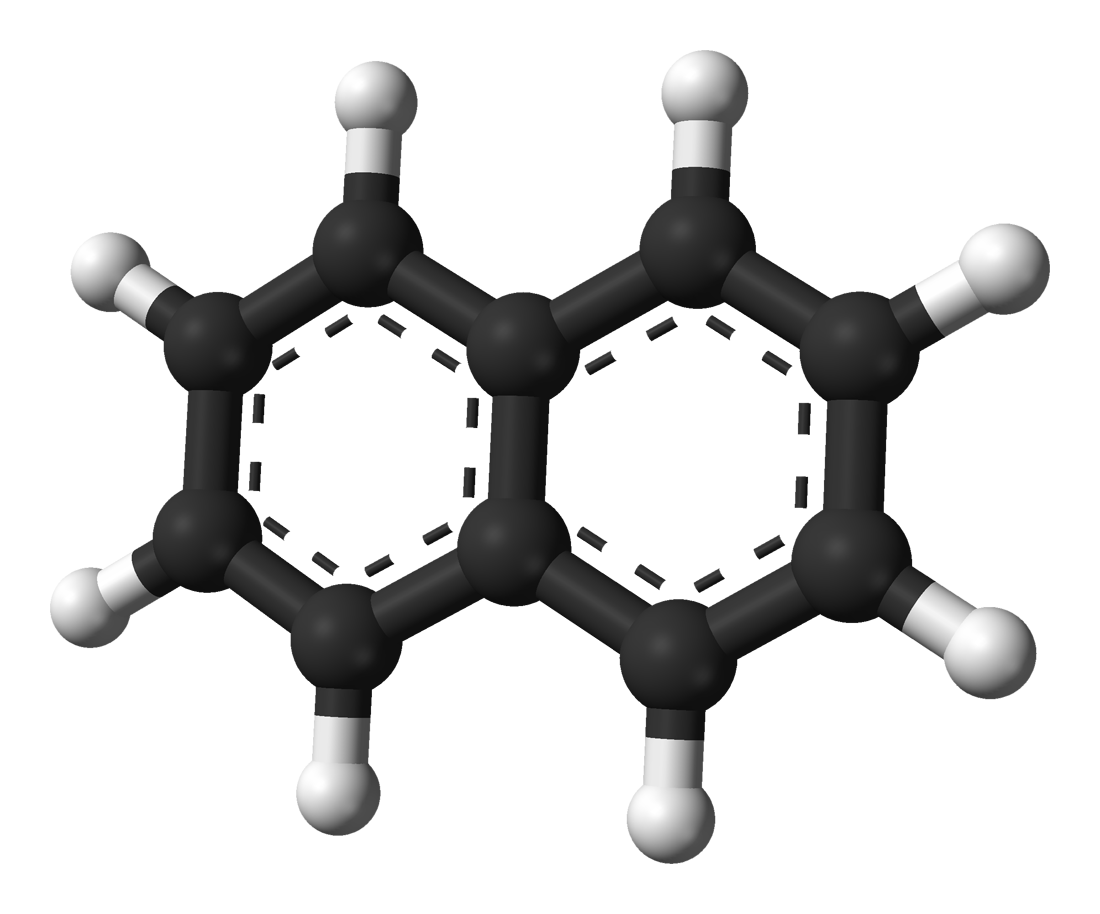}
    \caption{Naphthalane}
  \end{subfigure}
  \begin{subfigure}{0.24\textwidth}
    \centering
    \includegraphics[width=0.6\linewidth]{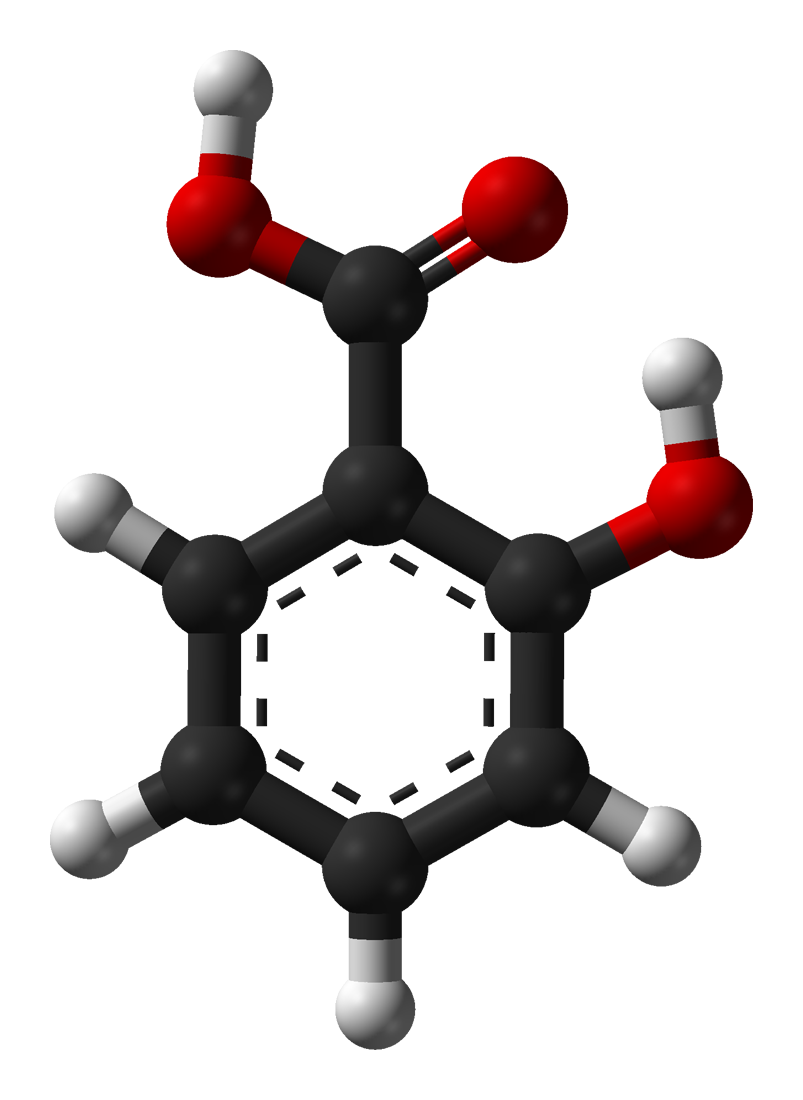}
    \caption{Salicyclic Acid}
  \end{subfigure}
  \begin{subfigure}{0.24\textwidth}
    \centering
    \includegraphics[width=0.6\linewidth]{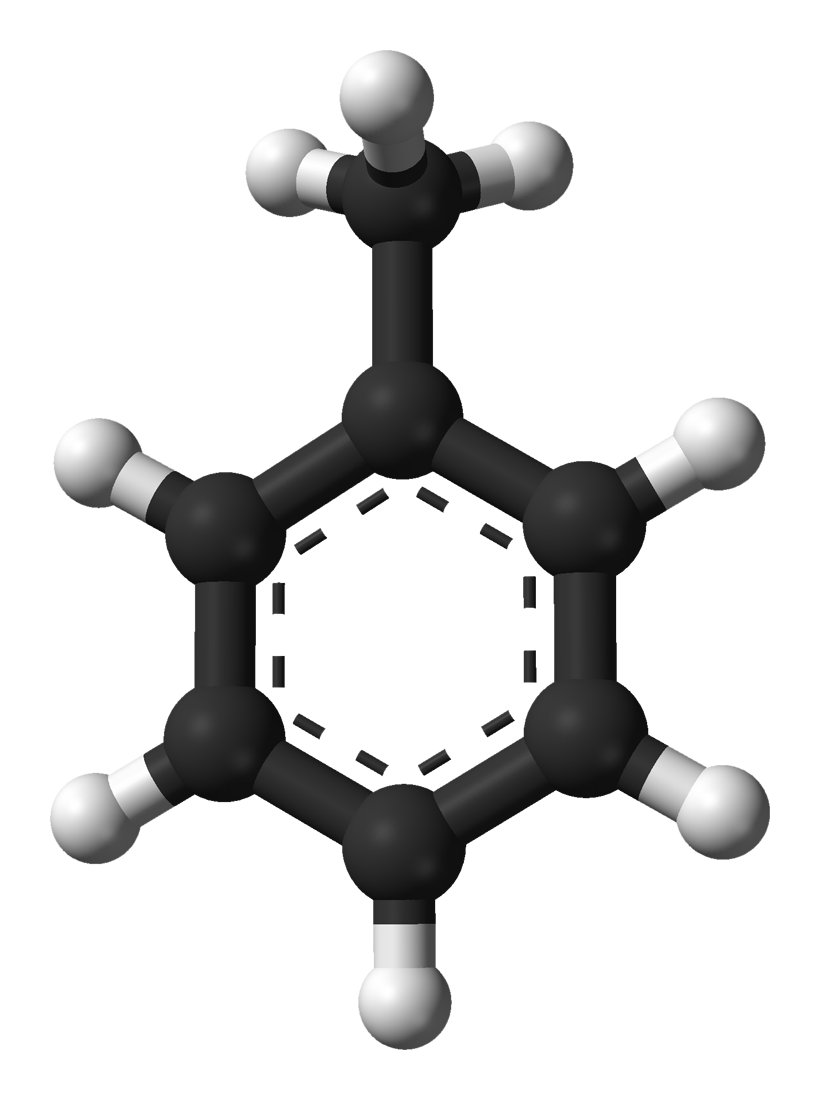}
    \caption{Toluene}
  \end{subfigure}
  \begin{subfigure}{0.25\textwidth}
    \centering
    \includegraphics[width=0.6\linewidth]{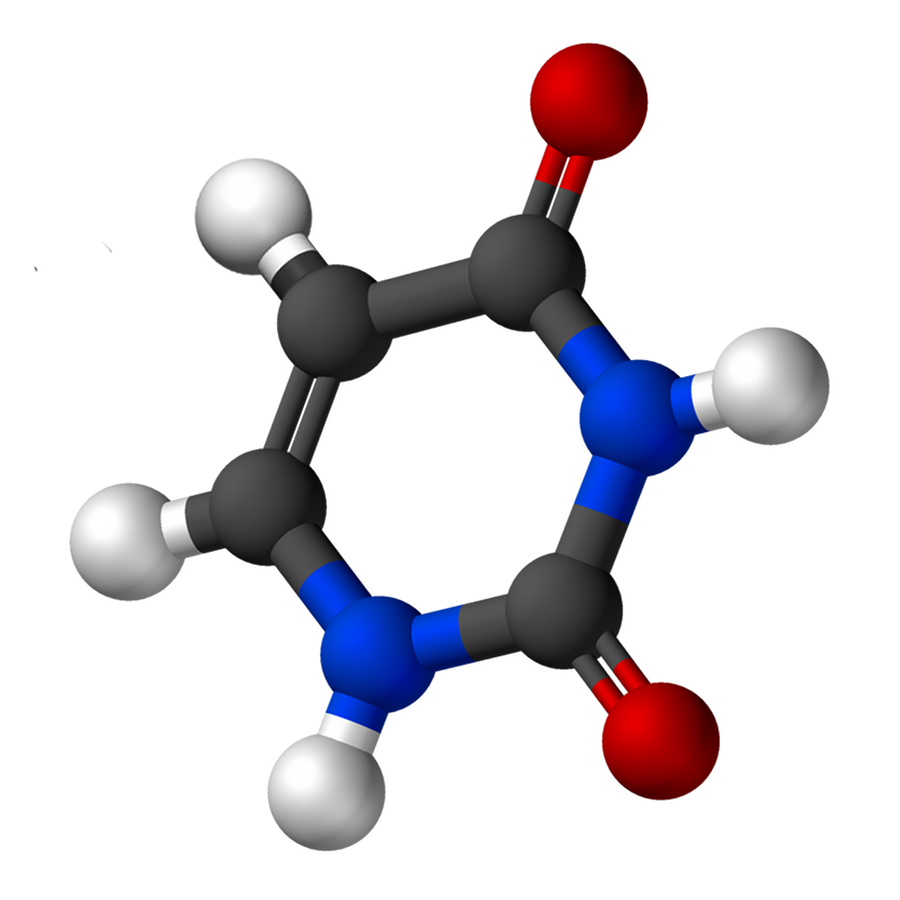}
    \caption{Uracil}
  \end{subfigure}

    \begin{tabular}{p{1.8in} p{1.1in} p{1.1in}}
    \toprule
       & \textbf{Training Set} \\
       Train on 1 molecule & f, b, a  \\
       Train on 2 molecules & ab, bc, de \\
       Train on 3 molecules & abg, abd, cef \\
       Train on 4 molecules & abeg, bcdf, abce \\
       Train on 5 molecules & abceg, abcde, bcdef \\
    \bottomrule
    \end{tabular}
    \end{center}
    \caption{Dataset Builds For Generalization Benchmarks}
    \label{table:exp2-sets}
\end{table}

First, we deterministically sample 1000 data points from each trajectory. These datasets are designated `a' to `g' and will be combined to create datasets that consist of multiple types of molecules. We select three different combinations of these datasets for one to five types of molecules. The dataset builds are shown in Table~\ref{table:exp2-sets}.
As shown in the table, the datasets are selected to create overlaps between combinations of the same number of molecule types (for example, abg vs. abd), and between the combinations of different numbers of molecule types (for example, ab vs. abg). This allows us to analyze the performance impact of progressively adding more trajectories to the datasets.

It is important to note that various molecular systems exhibit significantly different potential energy levels. When merging distinct trajectories into a single dataset, we normalize the energy levels by conducting a linear regression across the entire dataset to calculate the reference energy for each type of atom.

After constructing the fixtures, we proceed to train a model for each combined training set and individually evaluate their performance on the trajectories not present in the combined training set. We intentionally design the test set in this manner because when trajectory data for a known molecule is being tested, its performance will be significantly better than that of unseen molecules. Consequently, this would statistically obscure the performance patterns of the latter. We also remove molecules with a completely unseen species of atom (for example, Nitrogen in molecule g) in the training set from the test sets.

Furthermore, it is important to note that even if the partial energy contributions of all atoms are normalized by the reference energy points, the combined output should still be considered biased. This bias arises because the reference energy itself contains a high error margin, which would consequently offset the predicted potential energy level towards the most seen configurations. As a result, our primary focus lies on the forces where the offset bias is eliminated by the gradient operator.

\begin{figure}[t!]
  \centering
  \begin{subfigure}{0.48\textwidth}
    \centering
    \includegraphics[clip,width=1\linewidth]{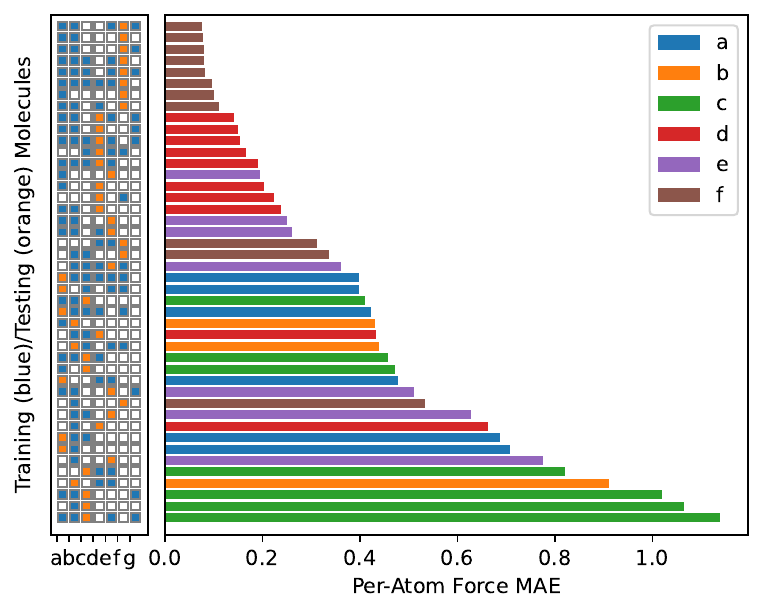}
    \caption{All Atoms}
  \end{subfigure}
  \begin{subfigure}{0.48\textwidth}
    \centering
    \includegraphics[clip, width=1\linewidth]{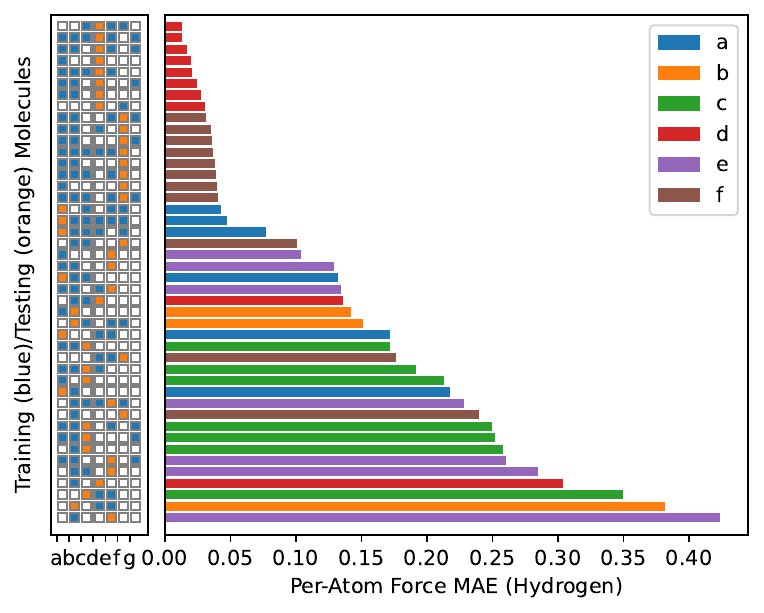}
    \caption{Hydrogen}
  \end{subfigure}

  \begin{subfigure}{0.48\textwidth}
    \centering
    \includegraphics[clip,width=1\linewidth]{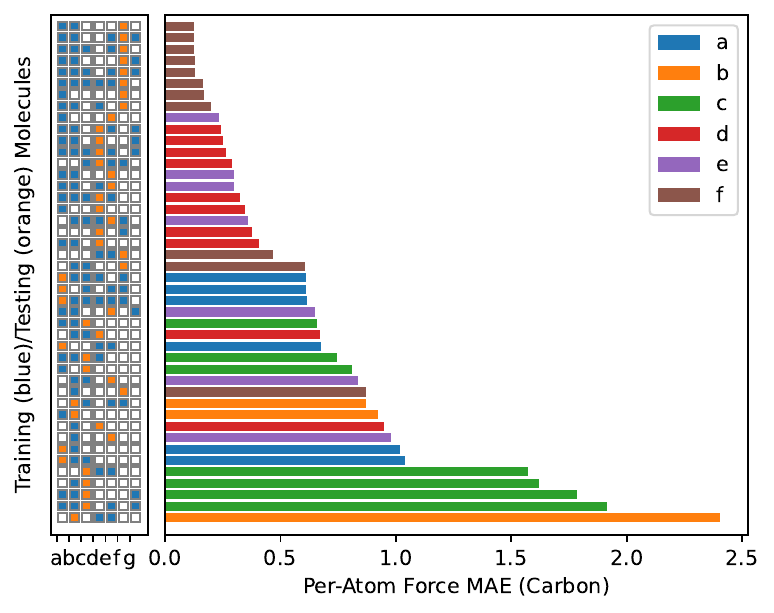}
    \caption{Carbon}
  \end{subfigure}
  \begin{subfigure}{0.48\textwidth}
    \centering
    \includegraphics[clip,width=1\linewidth]{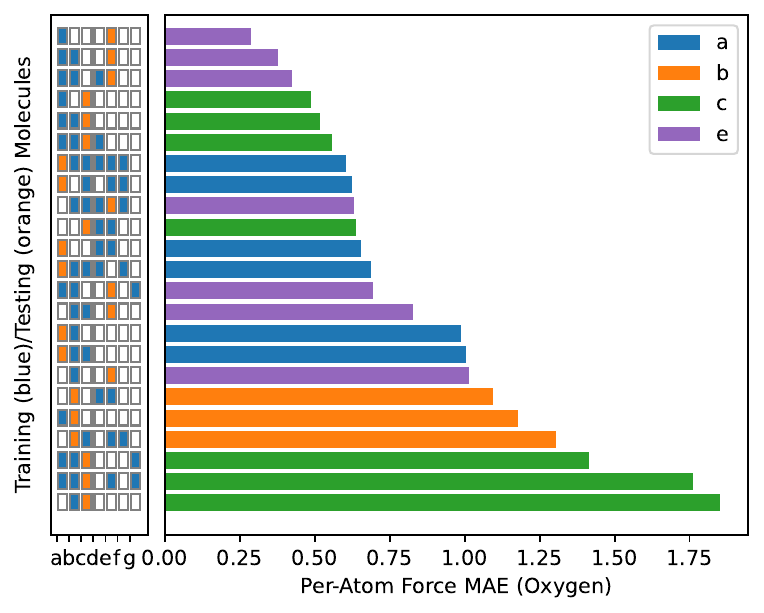}
    \caption{Oxygen}
  \end{subfigure}
  \caption{Cross-Molecule Generalization Benchmarks}
  \label{fig:3}
\end{figure}

Figure~\ref{fig:3} illustrates the cross-molecule generalization evaluation results. Similar to previous experiments, we present per-atom force MAE both over all the atoms and over different atom species. Each row in a plot represents training conducted on a specific combined dataset and testing conducted on one individual unseen molecule.
The left side of each plot presents the training (blue) and testing (orange) molecules, and the horizontal bar on the right presents the per-atom force MAE.

Upon sorting the rows by testing performance, a distinct pattern emerges, revealing that certain molecules (such as f and d) are easier to generalize than others, while some (like c) prove difficult to generalize. Intriguingly, molecules d and f are the only two that consist of just two types of atoms, namely Hydrogen and Carbon, while molecule c is the only one that contains two Oxygen atoms arranged in a symmetrical configuration.

The charts also reveal that different atom species do not consistently generalize best on the same molecule. While molecule f exhibits the overall best generalization, the top-performing molecule for each atom species varies: molecule d for Hydrogen, molecule f for Carbon, and molecule e for Oxygen.

This observation implies that the generalization performance is highly sensitive to the local structure and symmetries of the molecules. Although molecule d is the only one containing two Benzene rings and thus exhibits unique global structure features, its symmetries and the proximity around each atom resemble those of other molecules. Conversely, while the inter-atomic distances and dihedral angles in molecule c resemble other molecules, the symmetrical arrangement of dual Oxygen atoms results in configurations that are largely unseen in other molecules.

Counter-intuitively, expanding the training data to include a new molecule does not always enhance the model's ability to generalize onto a specific molecule. Rather, we observe that the quality of generalization largely depends on the similarity between the training and testing molecules. Expanding the training set to include a molecule with low similarity to the testing molecule will most likely decrease generalization performance.

This finding suggests that although neural networks excel at processing a large number of training samples with diverse characteristics, for the specific application of MD, if generalization over unseen molecules is a required capability, it may be more effective to consider the similarities of the molecules instead of feeding conflicting data into a single model. By partitioning the training data, training separate models, and dynamically routing the inference to the most compatible model, we can potentially achieve better generalization performance across a range of molecular structures.

\section{Conclusion}
Benchmarking on AI for Science requires careful design to combine the benchmarking steps for machine-learning-based AI methods and scientific computing. 
When the combined approach encounters conflicting assumptions, we override the conventional AI benchmarking settings with the scientific computing settings, for example, to embrace out-of-distribution problem instances.
The end result is an interleaved procedure that closely follows the conventional AI benchmarking practices by creating datasets, training AI models, and evaluating the performance on the test set but prioritizes scientifically meaningful setups in each step. More concretely, we demonstrate how this procedure is designed for MLFF-powered MD, a computational chemistry tool that plays a significant role in many scientific research applications.
Conventionally, MLFF evaluation adopts methodologies from AI benchmarking. This approach treats data points within the same trajectory as independent and identically distributed (i.i.d.), while data points in different trajectories of distinct molecules are considered to be completely differently distributed, often resulting in separate evaluations.
However, we argue that the evaluation of MLFFs should be tailored to accurately represent real-world MD computations.
This would enable the time-domain correlation of trajectory data to be exploited in order to assess the generalization capabilities of an MLFF model when predicting future time steps. Additionally, incorporating configurations from various molecules would test the model's adaptability to previously unseen spatial structures, which are common in chemical reaction simulations.
As a result, this leads to a realistic setup in the context of MLFF applications.
Taking advantage of the scientific research application setup, we can produce more scientifically meaningful performance metrics from the benchmark compared to conventional AI benchmarking methods, and contribute to the development of more robust and generalizable AI4S machine learning models.

\bibliographystyle{splncs04}
\bibliography{zotero}

\begin{thebibliography}{10}
\providecommand{\url}[1]{\texttt{#1}}
\providecommand{\urlprefix}{URL }
\providecommand{\doi}[1]{https://doi.org/#1}

\bibitem{bartokRepresentingChemicalEnvironments2013}
Bartók, A.P., Kondor, R., Csányi, G.: On representing chemical environments
  \textbf{87}(18),  184115. \doi{10.1103/PhysRevB.87.184115},
  \url{https://link.aps.org/doi/10.1103/PhysRevB.87.184115}

\bibitem{bartokGaussianApproximationPotentials2010}
Bartók, A.P., Payne, M.C., Kondor, R., Csányi, G.: Gaussian {{Approximation
  Potentials}}: {{The Accuracy}} of {{Quantum Mechanics}}, without the
  {{Electrons}}  \textbf{104}(13),  136403.
  \doi{10.1103/PhysRevLett.104.136403},
  \url{https://link.aps.org/doi/10.1103/PhysRevLett.104.136403}

\bibitem{batatiaMACEHigherOrder2022}
Batatia, I., Kovacs, D.P., Simm, G.N.C., Ortner, C., Csanyi, G.: {{MACE}}:
  {{Higher Order Equivariant Message Passing Neural Networks}} for {{Fast}} and
  {{Accurate Force Fields}}. \url{https://openreview.net/forum?id=YPpSngE-ZU}

\bibitem{batznerEquivariantGraphNeural2022}
Batzner, S., Musaelian, A., Sun, L., Geiger, M., Mailoa, J.P., Kornbluth, M.,
  Molinari, N., Smidt, T.E., Kozinsky, B.: E(3)-equivariant graph neural
  networks for data-efficient and accurate interatomic potentials
  \textbf{13}(1), ~2453. \doi{10.1038/s41467-022-29939-5},
  \url{https://www.nature.com/articles/s41467-022-29939-5}

\bibitem{behlerGeneralizedNeuralNetworkRepresentation2007}
Behler, J., Parrinello, M.: Generalized {{Neural-Network Representation}} of
  {{High-Dimensional Potential-Energy Surfaces}}  \textbf{98}(14),  146401.
  \doi{10.1103/PhysRevLett.98.146401},
  \url{https://link.aps.org/doi/10.1103/PhysRevLett.98.146401}

\bibitem{bischlASlibBenchmarkLibrary2016}
Bischl, B., Kerschke, P., Kotthoff, L., Lindauer, M., Malitsky, Y., Fréchette,
  A., Hoos, H., Hutter, F., Leyton-Brown, K., Tierney, K., Vanschoren, J.:
  {{ASlib}}: {{A}} benchmark library for algorithm selection  \textbf{237},
  41--58. \doi{10.1016/j.artint.2016.04.003},
  \url{https://www.sciencedirect.com/science/article/pii/S0004370216300388}

\bibitem{chmielaMachineLearningAccurate2017}
Chmiela, S., Tkatchenko, A., Sauceda, H.E., Poltavsky, I., Schütt, K.T.,
  Müller, K.R.: Machine learning of accurate energy-conserving molecular force
  fields  \textbf{3}(5),  e1603015. \doi{10.1126/sciadv.1603015},
  \url{https://www.science.org/doi/10.1126/sciadv.1603015}

\bibitem{christensenRoleGradientsMachine2020}
Christensen, A.S., von Lilienfeld, O.A.: On the role of gradients for machine
  learning of molecular energies and forces,
  \url{http://arxiv.org/abs/2007.09593}

\bibitem{dengImageNetLargescaleHierarchical2009}
Deng, J., Dong, W., Socher, R., Li, L.J., Li, K., Fei-Fei, L.: {{ImageNet}}:
  {{A}} large-scale hierarchical image database. In: 2009 {{IEEE Conference}}
  on {{Computer Vision}} and {{Pattern Recognition}}. pp. 248--255.
  \doi{10.1109/CVPR.2009.5206848}

\bibitem{fuForcesAreNot2022}
Fu, X., Wu, Z., Wang, W., Xie, T., Keten, S., Gomez-Bombarelli, R., Jaakkola,
  T.: Forces are not {{Enough}}: {{Benchmark}} and {{Critical Evaluation}} for
  {{Machine Learning Force Fields}} with {{Molecular Simulations}}.
  \doi{10.48550/arXiv.2210.07237}, \url{http://arxiv.org/abs/2210.07237}

\bibitem{gaoAIBenchScalableComprehensive2019}
Gao, W., Luo, C., Wang, L., Xiong, X., Chen, J., Hao, T., Jiang, Z., Fan, F.,
  Du, M., Huang, Y., Zhang, F., Wen, X., Zheng, C., He, X., Dai, J., Ye, H.,
  Cao, Z., Jia, Z., Zhan, K., Tang, H., Zheng, D., Xie, B., Li, W., Wang, X.,
  Zhan, J.: {{AIBench}}: {{Towards Scalable}} and {{Comprehensive Datacenter AI
  Benchmarking}}. In: Zheng, C., Zhan, J. (eds.) Benchmarking, {{Measuring}},
  and {{Optimizing}}, vol. 11459, pp.~3--9. {Springer International
  Publishing}. \doi{10.1007/978-3-030-32813-9_1},
  \url{http://link.springer.com/10.1007/978-3-030-32813-9_1}

\bibitem{gaoAIBenchIndustryStandard2019}
Gao, W., Tang, F., Wang, L., Zhan, J., Lan, C., Luo, C., Huang, Y., Zheng, C.,
  Dai, J., Cao, Z., Zheng, D., Tang, H., Zhan, K., Wang, B., Kong, D., Wu, T.,
  Yu, M., Tan, C., Li, H., Tian, X., Li, Y., Shao, J., Wang, Z., Wang, X., Ye,
  H.: {{AIBench}}: {{An Industry Standard Internet Service AI Benchmark
  Suite}}, \url{http://arxiv.org/abs/1908.08998}

\bibitem{gaoAIBenchScenarioScenarioDistilling2021}
Gao, W., Tang, F., Zhan, J., Wen, X., Wang, L., Cao, Z., Lan, C., Luo, C., Liu,
  X., Jiang, Z.: {{AIBench Scenario}}: {{Scenario-Distilling AI Benchmarking}}.
  In: 2021 30th {{International Conference}} on {{Parallel Architectures}} and
  {{Compilation Techniques}} ({{PACT}}). pp. 142--158. {IEEE}.
  \doi{10.1109/PACT52795.2021.00018},
  \url{https://ieeexplore.ieee.org/document/9563026/}

\bibitem{jiaPushingLimitMolecular2020}
Jia, W., Wang, H., Chen, M., Lu, D., Lin, L., Car, R., E, W., Zhang, L.:
  Pushing the limit of molecular dynamics with ab initio accuracy to 100
  million atoms with machine learning, \url{http://arxiv.org/abs/2005.00223}

\bibitem{jonesDeterminationMolecularFields1997}
Jones, J.E., Chapman, S.: On the determination of molecular fields.—{{I}}.
  {{From}} the variation of the viscosity of a gas with temperature
  \textbf{106}(738),  441--462. \doi{10.1098/rspa.1924.0081},
  \url{https://royalsocietypublishing.org/doi/10.1098/rspa.1924.0081}

\bibitem{argonnenationallaboratoryAIScienceReport}
Laboratory, A.N.: {{AI}} for {{Science Report}},
  \url{https://publications.anl.gov/anlpubs/2020/03/158802.pdf}

\bibitem{mattsonMLPerfTrainingBenchmark}
Mattson, P., Cheng, C., Coleman, C., Diamos, G., Micikevicius, P., Patterson,
  D., Tang, H., Wei, G.Y., Bailis, P., Bittorf, V., Brooks, D., Chen, D.,
  Dutta, D., Gupta, U., Hazelwood, K., Hock, A., Huang, X., Ike, A., Jia, B.,
  Kang, D., Kanter, D., Kumar, N., Liao, J., Ma, G., Narayanan, D., Oguntebi,
  T., Pekhimenko, G., Pentecost, L., Reddi, V.J., Robie, T., John, T.S.,
  Tabaru, T., Wu, C.J., Xu, L., Yamazaki, M., Young, C., Zaharia, M.: {{MLPerf
  Training Benchmark}} p.~14

\bibitem{musaelianLearningLocalEquivariant2023}
Musaelian, A., Batzner, S., Johansson, A., Sun, L., Owen, C.J., Kornbluth, M.,
  Kozinsky, B.: Learning local equivariant representations for large-scale
  atomistic dynamics  \textbf{14}(1), ~579. \doi{10.1038/s41467-023-36329-y},
  \url{https://www.nature.com/articles/s41467-023-36329-y}

\bibitem{thiyagalingamScientificMachineLearning2022}
Thiyagalingam, J., Shankar, M., Fox, G., Hey, T.: Scientific machine learning
  benchmarks  \textbf{4}(6),  413--420. \doi{10.1038/s42254-022-00441-7},
  \url{https://www.nature.com/articles/s42254-022-00441-7}

\bibitem{wangBigDataBenchBigData2014}
Wang, L., Zhan, J., Luo, C., Zhu, Y., Yang, Q., He, Y., Gao, W., Jia, Z., Shi,
  Y., Zhang, S., Zheng, C., Lu, G., Zhan, K., Li, X., Qiu, B.:
  {{BigDataBench}}: {{A}} big data benchmark suite from internet services. In:
  2014 {{IEEE}} 20th {{International Symposium}} on {{High Performance Computer
  Architecture}} ({{HPCA}}). pp. 488--499. \doi{10.1109/HPCA.2014.6835958}

\bibitem{wangViSNetScalableAccurate2022}
Wang, Y., Li, S., He, X., Li, M., Wang, Z., Zheng, N., Shao, B., Wang, T., Liu,
  T.Y.: {{ViSNet}}: A scalable and accurate geometric deep learning potential
  for molecular dynamics simulation, \url{https://arxiv.org/abs/2210.16518v1}

\bibitem{wangImprovingMachineLearning2023}
Wang, Z., Wu, H., Sun, L., He, X., Liu, Z., Shao, B., Wang, T., Liu, T.Y.:
  Improving machine learning force fields for molecular dynamics simulations
  with fine-grained force metrics  \textbf{159}(3),  035101.
  \doi{10.1063/5.0147023}

\bibitem{zhangArtificialIntelligenceScience2023}
Zhang, X., Wang, L., Helwig, J., Luo, Y., Fu, C., Xie, Y., Liu, M., Lin, Y.,
  Xu, Z., Yan, K., Adams, K., Weiler, M., Li, X., Fu, T., Wang, Y., Yu, H.,
  Xie, Y., Fu, X., Strasser, A., Xu, S., Liu, Y., Du, Y., Saxton, A., Ling, H.,
  Lawrence, H., Stärk, H., Gui, S., Edwards, C., Gao, N., Ladera, A., Wu, T.,
  Hofgard, E.F., Tehrani, A.M., Wang, R., Daigavane, A., Bohde, M., Kurtin, J.,
  Huang, Q., Phung, T., Xu, M., Joshi, C.K., Mathis, S.V., Azizzadenesheli, K.,
  Fang, A., Aspuru-Guzik, A., Bekkers, E., Bronstein, M., Zitnik, M.,
  Anandkumar, A., Ermon, S., Liò, P., Yu, R., Günnemann, S., Leskovec, J.,
  Ji, H., Sun, J., Barzilay, R., Jaakkola, T., Coley, C.W., Qian, X., Qian, X.,
  Smidt, T., Ji, S.: Artificial {{Intelligence}} for {{Science}} in
  {{Quantum}}, {{Atomistic}}, and {{Continuum Systems}}.
  \doi{10.48550/arXiv.2307.08423}, \url{http://arxiv.org/abs/2307.08423}

\end{thebibliography}

\end{document}